# Multiple models of Bayesian networks applied to offline recognition of Arabic handwritten city names

**Mohamed Ali Mahjoub, Nabil Ghanmy, Khlifia jayech** and **Ikram Miled**

Research Unit SAGE, National Engineering School of Sousse
Technology Park, 4054, Sahloul, Sousse, Tunisia
Medali.mahjoub@ipeim.rnu.tn, Nabil.ghanmy@gmail.com, Jayech_k@yahoo.f,Miledikram@yahoo.fr

**ABSTRACT**

*In this paper we address the problem of offline Arabic handwriting word recognition. Off-line recognition of handwritten words is a difficult task due to the high variability and uncertainty of human writing. The majority of the recent systems are constrained by the size of the lexicon to deal with and the number of writers. In this paper, we propose an approach for multi-writers Arabic handwritten words recognition using multiple Bayesian networks. First, we cut the image in several blocks. For each block, we compute a vector of descriptors. Then, we use K-means to cluster the low-level features including Zernik and Hu moments. Finally, we apply four variants of Bayesian networks classifiers (Naïve Bayes, Tree Augmented Naïve Bayes (TAN), Forest Augmented Naïve Bayes (FAN) and DBN (dynamic bayesian network) to classify the whole image of tunisian city name. The results demonstrate FAN and DBN outperform good recognition rates.*

## 1. INTRODUCTION

Machine perception and recognition of handwritten text in any language is a difficult problem. In this context, segmentation and recognition of Arabic handwritten words are two important areas of research (S Touj et al 2007). It has many applications in bank check reading, mail sorting, forms processing in administration and insurance. Arabic script presents additional challenges for handwriting recognition systems due to its highly connected nature, numerous forms of each letter, presence of ligatures, and regional differences in writing styles and habits. Off-line Arabic Handwriting Recognition is a difficult task due to the high variability and uncertainty of human writing (infinite variations of shapes resulting from the writing habit, scanning methods, fusion of diacritical points, overlapping …).

The field of handwriting recognition can be split into two different approaches; on line recognition and offline recognition (Miled H. 1998, Mantas J. 1991). The first approach deals with the recognition of handwriting captured by a tablet or similar device, and uses the digitised trace of the pen to recognise the symbol. In this instance the recogniser will have access to the x and y coordinates as a function of time, and thus has temporal information about how the symbol was formed. The second approach concentrates on the recognition of handwriting in the form of an image. In this instance only the completed character or word is available. It is this off-line approach that will be taken in this report. Many works proposed different approaches such as statistical, structural, neural, and Markov, but all these approaches have many limits. There are two questions to be answered in order to solve difficulties that are hampering the progress of research in this direction. First, how to link semantically objects in images with high-level features? That’s mean how to learn the dependence between objects that reflected better the data? Secondly, how to classify the image using the found dependencies structure? Our paper presents a work which uses four variants of Bayesian network to classify image of handwritten Arabic words using the structure of dependence finding between objects; classifiers (Naïve Bayes, Tree Augmented Naïve Bayes (TAN), Forest Augmented Naïve Bayes (FAN) and DBN (dynamic bayesian network). A dynamic Bayesian network can be defined as a repetition of conventional networks in which we add a causal one time step to another. Each Network contains a number of random variables representing observations and hidden states of the process. In our model, the DBN is considered as coupling two HMMs. The main difference between the HMM and dynamic Bayesian networks is that in an DBN the hidden states are represented as distributed by a set of random variables $(X_t^1, X_t^2, \ldots . X_t^n)$. Thus, in an HMM, the state space consists of a single random variable $X_t$. DBNs are another extension of HMMs which have been recently applied to speech recognition.

This paper is divided as follows: Section 2 presents an overview of main researches dealing with offline recognition of Arabic handwritten characters. Section 3 provides a reminder of the basics of Bayesian networks. Section 4 presents methods of feature extraction. Sections 5 and 6 describe the four approaches where we introduce the method of building the Naïve Bayesian Network and Dynamic Bayesian Network. In Section 7, we present some experiments, results and discussions; finally, Section 8 presents the conclusion.

## 2. OFFLINE ARABIC WORD RECOGNITION

During the last several years, off-line Arabic handwriting recognition has been a popular topic in the OCR/ICR research community. According to the recent publications, the research and development efforts are growing leading to important advances in this area. Like on line recognition (Mahjoub, 1999), the off-line handwriting recognition, represents the most challenging OCR problem because of the different variations of writings and the absence of dynamic information relative to the different considered shapes (Bozinovik & Srihari, 1989; N. Ben Amara 2000). In order to resolve these problems, many approaches have been proposed. The majority of the developed systems have been inspired by works on human reading modes (Cote, 1997). In this context, several classifiers have been tested such as 1D HMMs with different topologies corresponding to different levels of perception in the works of Miled (Miled, 1998). Transparent neural network classifiers were proposed in (Souici-Meslati et al., 2002 ; Maddouri et al., 2002). Analytic approaches have also been adopted in the works of Atici (Pechwitz et al., 2001) and Pechwitz (Pechwitz & Margner, 2003). Different criteria can condition the complexity of handwriting OCR systems (infinite variations of shapes resulting from the writing habit, style and other conditions of the writer as well as other factors such as the writing instrument, writing surface, scanning methods (Suen, 1973), fusion of diacritical points, overlapping and touching of sub-words...and of course the machine’s character recognition algorithms (Mantas, 1991). Following, we will focus in particular on problems related to the intrinsic writer's variations (figure 1). These problems are generally the most complex, especially in the case of Arabic. According to the conducted study, the important problems are related to writing discontinuity and slant, overlapping and sub-words touching, shape discrimination and variations in subword sizes. The diacritical marks cause serious problems.

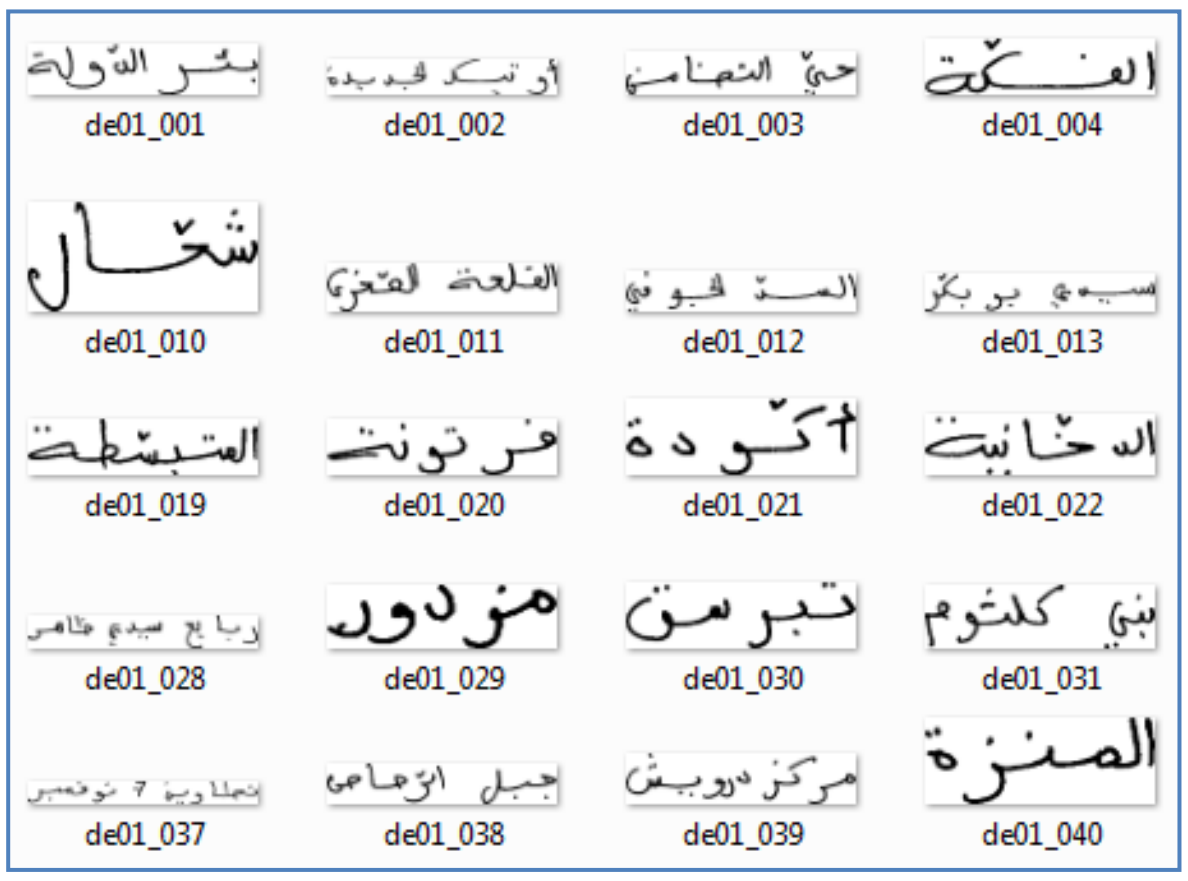

**Figure 1:** Town names data base

## 3. BAYESIAN NETWOK

### 3.1 Bayesian net

Bayesian networks represent a set of variables in the form of nodes on a directed acyclic graph (DAG). It maps the conditional independencies of these variables (D. Heckerman, 1995). Bayesian networks bring us four advantages as a data modeling tool Firstly, Bayesian networks are able to handle incomplete or noisy data which is very frequently in image analysis (D. Bellot 2002). Secondly, Bayesian networks are able to ascertain causal relationships through conditional independencies, allowing the modeling of relationships between variables. The last advantage is that Bayesian networks are able to incorporate existing knowledge, or pre-known data into its learning, allowing more accurate results (Friedman et al. 1996) by using what we already know.

Bayesian network is defined by:

- A directed acyclic graph (DAG) G= (V, E), where V is a set of nodes of G, and E of of G ;
- A finite probabilistic space $(\Omega, Z, p)$ ;
- A set of random variables associated with graph nodes and defined on $(\Omega, Z, p)$ as :

$$p(V_1, V_2, \dots, V_n) = \prod_{i=1}^{n} p(V_i | C(V_i))$$

where $C(V_i)$ is a set causes (parents) $V_i$ in graph G.

A Bayesian network is then composed of two components; a causal directed acyclic graph (qualitative representation of knowledge) and a set of local distributions of probability. After defining the different values that can take a characteristic, the expert must indicate the different relationships between these characteristics. Finally, the definition of the different probabilities is necessary to complete the network, each value (state) of a given node must have the probability of occurrence.

Suppose we have a Bayesian network defined by a graph and the probability distribution associated with (G, P). Suppose that the graph is composed of n nodes, denoted $X = \{X_1, X_2, \ldots, X_n\}$. The general problem of inference is to compute $p(X_i|Y)$ where $Y \subset X$ and $X_i \notin Y$. To calculate these conditional probabilities we can use methods of exact or approximate inferences. The first gives an accurate result, but is extremely costly in time and memory. The second turn requires less resources but the result is an approximation of the exact solution. A BN is usually transformed into a decomposable Markov network for inference. During this transformation, two graphical operations are performed on the DAG of a BN, namely, moralization and triangulation.

### 3.2 Learning Bayesian network

There are several problems in the use of BNs, we will cite the mains (Pechwitz et al., 2001). The first one; The correspondence between the graphical structure and associated probabilistic structure will allow to reduce all the problems of inference problems in graph theory. However, these problems are relatively complex and give rise to much research. The second difficulty of Bayesian networks lies precisely in the operation for transposition of the causal graph to a probabilistic representation. Even if the only probability tables needed to finish the entire probability distribution are those of a node conditioning compared to his parents, he is the definition of these tables is not always easy for an expert. Another problem of Bayesian networks, the problem of automatic learning of the structure that remains is a rather complex problem.

In this case the structure is completely known a priori and all variables are observable from the data, the learning of conditional probabilities associated with variables (network nodes) may be from either a randomly or according to a Bayesian approach (J. Pearl 1988). The statistical learning calculation value frequencies in the training data is based on the maximum likelihood (ML). Structure learning is the act of finding a plausible structure for a graph based on data input. However, it has been proven that this is an NP-Hard problem, and therefore any learning algorithm that would be appropriate for use on such a large dataset such as microarray data would require some form of modification for it to be feasible. It is explained confirmed that finding the most appropriate DAG from sample data is large problem as the number of possible DAGs grows super-exponentially with the number of nodes present. In practice we use heuristics and approximation methods like K2 algorithm (M. Deviren et al 2001).

A variant of Bayesian Network is called Naïve Bayes. Naïve Bayes is one of the most effective and efficient classification algorithms**.** The conditional independence assumption in naïve Bayes is rarely true in reality. Indeed, naive Bayes has been found to work poorly for regression problems, and produces poor probability estimates. One way to alleviate the conditional independence assumption is to extend the structure of naive Bayes to represent explicitly attribute dependencies by adding arcs between attributes.

# 4. FEATURE EXTRACTION

## 4.1 Feature extraction and principal component analysis

Each block is described by a feature vector. This vector is obtained by combining the Zernike moments and Hu. In general, a single descriptor does not give satisfactory recognition rate. One solution is to combine different descriptors. This solution seems to be more robust than using a single descriptor. We present the descriptors we chose to combine. We calculated the Zernike moments of order 0 to order 2 based on the formulas detailed in Chapter 1 section 2.3.1. Finally we divide by the Zernike moments of time Z (0.0) that they belong to the interval [0,1]. Hu moments are calculated from normalized moments and are invariant under translation, rotation and scaling. Hu moments are calculated from normalized moments and are invariant under translation, rotation and scaling. Thus, these descriptors are used as input vectors for training and testing of a Bayesian network. The feature vector is constructed of 7 Hu invariants descriptors and 5 invariant Zernike descriptors. Their concatenation gives us a vector of 12 components per image block. Descriptors described above provide us with signatures of continuous values. However, the naive Bayesian classifier requires discrete variables. To overcome this problem, we used a discretization method to transform the variables with continuous values into variables with discrete values

The Principal Component Analysis (PCA) is a method of family data analysis and more generally of multivariate statistics, which is to transform these variables (called "correlated" in statistics) to new independent variables each other (thus "uncorrelated"). These new variables are called "principal components", or axes. It allows the practitioner to reduce the information in a more limited number of components as the initial number of variables. From a set of N images in a space of p descriptors, the goal of this method is to find a representation in a small space of q dimensions (q <<p) which preserves the "best summary" (as defined the maximum projected variance).

- Input : All continuous variables structured in a matrix M with N rows and P column.

$$M = \begin{bmatrix} X_{11} & \cdots & X_{1P} \\ \vdots & \ddots & \vdots \\ X_{N1} & \cdots & X_{NP} \end{bmatrix}$$

- Output : variables *:* $C_1, C_2, \ldots, C_k, \ldots, C_q$ where $q < p$ *and* $C_k$ are*:*
    - 2-2 uncorrelated,
    - of maximum variance
    - of decreasing importance

Algorithm

1- focus and reduce the variables. Thus, each random variable $X_i = (X_{1i}, \ldots, X_{Ni})'$ *has an overage* $\bar{X}_i$ and standard deviation $\sigma_{X_i}$.

$$\widetilde{M} = \begin{bmatrix} \frac{X_{11} - \bar{X}_1}{\sigma(X_1)} & \cdots & \frac{X_{1P} - \bar{X}_P}{\sigma(X_P)} \\ \vdots & \ddots & \vdots \\ \frac{X_{N1} - \bar{X}_N}{\sigma(X_N)} & \cdots & \frac{X_{NP} - \bar{X}_P}{\sigma(X_P)} \end{bmatrix}$$

2- Determine the correlation matrix to measure the dependence between variable.

The maximum number of principal components it is possible to extract a correlation matrix equals the number of variables in the matrix. In our case so we could extract up to 35 components. We analyze the values to determine how many components are worth extracting. This is well detailed in the following section.

### 4.2 Discretization method

Our goal is to assign each continuous variable to a class using the K-means algorithm. K-means clustering is a partitioning method. The function k-means partitions data into k mutually exclusive clusters, and returns the index of the cluster to which it has assigned each observation. It treats each observation in your data as an object having a location in space. It finds a partition in which objects within each cluster are as close to each other as possible, and as far from objects in other clusters as possible. If we know the number of clusters, we can call k-means with K, the desired number. K-means clustering is a partitioning method. The function k-means partitions data into k mutually exclusive clusters, and returns the index of the cluster to which it has assigned each observation. It treats each observation in your data as an object having a location in space. It finds a partition in which objects within each cluster are as close to each other as possible, and as far from objects in other clusters as possible. If you know the number of clusters, we can call k-means with K, the desired number: $\mathrm{IDX} = \mathrm{Kmeans}\ (\mathrm{X}, \mathrm{K})$. The problem appears when you don't know the optimal value of K. To get an idea of how well-separated the resulting clusters are, you can make a silhouette plot using the cluster indices output from k-means. The silhouette plot displays a measure of how close each point in one cluster is to points in the neighboring clusters. This measure ranges from +1, indicating points that are very distant from neighboring clusters, through 0, indicating points that are not distinctly in one cluster or another, to -1, indicating points that are probably assigned to the wrong cluster. According to our tests, *silhouette* returns these values in its first output: $\mathrm{IDX} = \mathrm{Kmeans}\ (\mathrm{X}, 21)$; Silhouette $(X, IDX)$

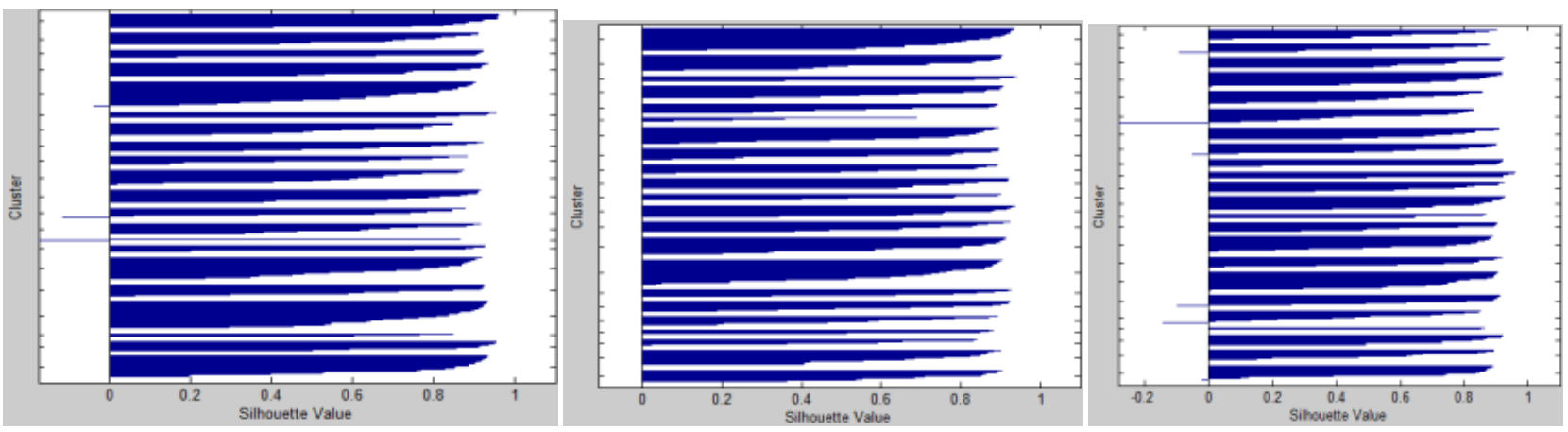

a) K=20 b) k=22 c)k=23

**Figure 2.** Silhouette value (number of cluster k)

From the silhouette (figure 2) plot, we can see that some clusters are separated from neighboring clusters. However, there are 3 clusters that contain a few points with negative values, indicating that those clusters are not well separated. For this reason, we increase the number of clusters to see if k-means can find a better grouping of the data. So, we choose as value of K = 22. A silhouette plot for this solution indicates that these twenty-two clusters are better separated than the twenty-one in the previous solution. Finally, we will try clustering the data using twenty-three clusters (figure 2). This silhouette plot indicates that this is probably not the right number of clusters, since five of the clusters contain points with mostly low silhouette values. Without some knowledge of how many clusters are really in the data, it is a good idea to experiment with a range of values for k.

## 5. NAÏVE BAYESIAN NETWORK

The proposed system comprises three main modules: a module for extracting primitive blocks from the local cutting of words images, a classification module of primitives in the cluster by using the method of k-means, and a classification module the overall picture based on the structures of Bayesian networks developed for each class. The developed classification system is illustrated in figure 3. The following sub-sections describe the different step of the proposed system.

### 5.1 Step 1: decomposition of images into blocks

Classifying an image depends on classifying the objects within the image. The studied objects in an image are the following analysis represented by blocks. An example of one image being divided into elementary building blocks that contain different concepts is illustrated in figure 4.

### 5.2 Step 2: Features Extraction

After the pretreatment of the word, we use the descriptors such as moment invariants of Zernike and Hu descriptors, which are invariant to rotation, translation and scaling, to extract the characteristics of each block.

**Zernike descriptor:** They are constructed using a set of complex polynomials which form a complete orthogonal set on the unit disk with $(\mathbf{x^2} + \mathbf{y^2}) < \mathbf{1}$ :

$$A_{mn} = \frac{m+1}{\pi} \iint I(x, y)\,[V_{mn}(x, y)]\,dx\,dy$$

Where *m* and *n* define the order of the moment and *I (x, y)* the gray level of a pixel in the image. The Zernike polynomials $\boldsymbol{V_{mn}(x, y)}$ are expressed in polar coordinates:

$$V_{mn}(r, \theta) = R_{mn}(r)e^{-jn\theta}$$

Where $\boldsymbol{R_{mn}(r)}$ is the radial orthogonal polynomial:

$$R_{mn}(r) = \sum_{s=0}^{\frac{m-|n|}{2}} (-1)^s \frac{(m-s)!}{s!\left(\frac{m+|n|}{2} - s\right)!\left(\frac{m-|n|}{2} - s\right)!} r^{m-2s}$$

**Hu descriptor:** Hu made the first step in the use of moment invariants for pattern recognition by offering the seven Hu moments which are calculated from the normalized moments. Hu moments are invariant to translation, rotation and scaling. We introduce the normalized moments as follows: $\mathbf{u_{p,q}} = \frac{v_{p,q}}{v_{0,0}^{1+(p+q)/2}}$. The seven moments of Hu are:

$$\varphi_1 = u_{2,0} + u_{0,2}$$
$$\varphi_2 = (u_{2,0} - u_{0,2})^2 + 4u_{1,1}^2$$
$$\varphi_3 = (u_{3,0} - 3u_{1,2})^2 + (3u_{2,1} - u_{0,3})^2$$
$$\varphi_4 = (u_{3,0} + u_{1,2})^2 + (u_{2,1} + u_{0,3})^2$$
$$\varphi_5 = (u_{3,0} - 3u_{1,2})(u_{3,0} + u_{1,2})\left[(u_{3,0} + u_{1,2})^2 - 3(u_{2,1} + u_{0,3})^2\right] + (3u_{2,1} - u_{0,3})(u_{2,1} + u_{0,3})\left[3(u_{3,0} + u_{1,2})^2 - (u_{2,1} + u_{0,3})^2\right]$$
$$\varphi_6 = (u_{2,0} - u_{0,2})\left[(u_{3,0} + u_{1,2})^2 - (u_{2,1} + u_{0,3})^2\right] + 4u_{1,1}(u_{3,0} + u_{1,2})(u_{2,1} + u_{0,3})$$
$$\varphi_7 = (3u_{2,1} - u_{0,3})(u_{3,0} + u_{1,2})\left[(u_{3,0} + u_{1,2})^2 - 3(u_{2,1} + u_{0,3})^2\right] - (u_{3,0} - 3u_{1,2})(u_{2,1} + u_{0,3})\left[3(u_{3,0} + u_{1,2})^2 - (u_{2,1} + u_{0,3})^2\right]$$

### 5.3 Step 3: Clustering of blocs with K-means

Our approach is based on modeling of the image by three blocks reflecting a local description of the image (figure 4). At each block we will first apply the descriptors presented in the previous section and second we classify it by Bayesian classifier. To generate the label vector from vector descriptor, we used the k-means algorithm. Each vector will undergo a clustering attribute, and replace the labels generated vector components descriptors. We use the method of k-means to cluster the descriptor as shown in figure 4.

### 5.4 Step 4: Structure Learning

The originality of this work is the development of a Bayesian network for each block. Then, each word image is characterized by three Bayesian networks. We propose three variants of Bayesian Networks such as Naïve Bayesian Network (NB), Tree Augmented Naïve Bayes (TAN) and Forest Augmented Naïve Bayes (FAN). For more details see references (Jayech & Mahjoub, 2010 ; Jayech & Mahjoub, 2011).

### 5.5 Step 5: Parameters learning

NB, TAN and FAN classifiers parameters were obtained by using the procedure described in (Jayech & Mahjoub, 2010; Jayech & Mahjoub, 2011). In implementation of NB, TAN and FAN, we used the Laplace estimation to avoid the zero-frequency problem. More precisely, we estimated the probabilities $P(c)$, $P(a_i|c)$ and $P(a_i|a_j, c)$ using Laplace estimation (Miled, 1998).

### 5.6 Step 6: Classification

In this work the decisions are inferred using Bayesian Networks. Class of an example is decided by calculating posterior probabilities of classes using Bayes rule. This is described for both classifiers. We suppose that $C^i$ class is composed with three subclasses $C_t^i$ linked in over time (t=1,2,3). One image is divided on three blocks. We start to classify the blocks $B_k$ with their attributes using Bayes rules :

- The blocks $\mathrm{B_{k1}}$ at time t=1 considering $C_1^i$ classes :

$$p(C_1^i|B_{k1}) = p(C_1^i|A_{1b_1},\dots,A_{mb_1}) = \frac{p(A_{1b_1},\dots,A_{mb_1}|C_1^i)*p(C_1^i)}{p(A_{1b_1},\dots,A_{mb_1})} = \frac{\prod_j p(A_{jb_1}|C_1^i)*p(C_1^i)}{p(A_{1b_1},\dots,A_{mb_1})}$$

- The blocks $\mathrm{B_{k2}}$ at time t=2 considering $C_2^i$ classes :

$$p(C_2^i|B_{k2}) = p(C_2^i|A_{1b_2},\dots,A_{mb_2}) = \frac{p(A_{1b_2},\dots,A_{mb_2}|C_2^i)*p(C_2^i)}{p(A_{1b_2},\dots,A_{mb_2})} = \frac{\prod_j p(A_{jb_2}|C_2^i)*p(C_2^i)}{p(A_{1b_2},\dots,A_{mb_2})}$$

- The blocks $\mathrm{B_{k2}}$ at time t=3 considering $C_3^i$ classes :

$$p(C_3^i|B_{k3}) = p(C_3^i|A_{1b_3},\dots,A_{mb_3}) = \frac{p(A_{1b_3},\dots,A_{mb_3}|C_3^i)*p(C_3^i)}{p(A_{1b_3},\dots,A_{mb_3})} = \frac{\prod_j p(A_{jb_3}|C_3^i)*p(C_3^i)}{p(A_{1b_3},\dots,A_{mb_3})}$$

Note that we do not need to explicitly calculate the denominators:

$$p(A_{1b_1},\dots,A_{mb_1}), p(A_{1b_3},\dots,A_{mb_3}), p(A_{1b_2},\dots,A_{mb_2})$$

They are determinate by the normalization condition. Thus it's sufficient to calculate for each sub class $\mathrm{C_t^i}$ its likelihood degree:

$$\prod_j p(A_{jb_t}|C_t^i)*p(C_t^i)$$

We move to classify the whole image from its entire three blocks considering $C^i$ classes:

$$p(C^i|image_k) = p(C^i|B_{k1},B_{k2},B_{k3}) = \frac{1}{3}\sum_{t=3}^{3} p(C_t^i|B_{kt})$$

To determine the subclass of each block the Bayesian approach do the comparison between the conditional probability $p(C_t^i|B_{kt})$ for each time t and chose the maximum value by applying the Bayes decision rule as follows :

$$d(x) = \mathrm{argmax}_{C_t^i}\, p(C_t^i|B_{kt}) = \mathrm{argmax}_{C_t^i} \prod_j p(A_{jb_t}|C_t^i).\,p(C_t^i)$$

To find the image class considering $\mathrm{C^i}$ classes the Bayesian method do the comparison between the conditional probabilities $p(C^i|image_j)$.

$$d(x) = \mathrm{argmax}_{C^i}\, p(C^i|image_k) = \mathrm{argmax}_{C^i} \frac{1}{3}\sum_{t=1}^{3} p(C_t^i|B_{kt})$$

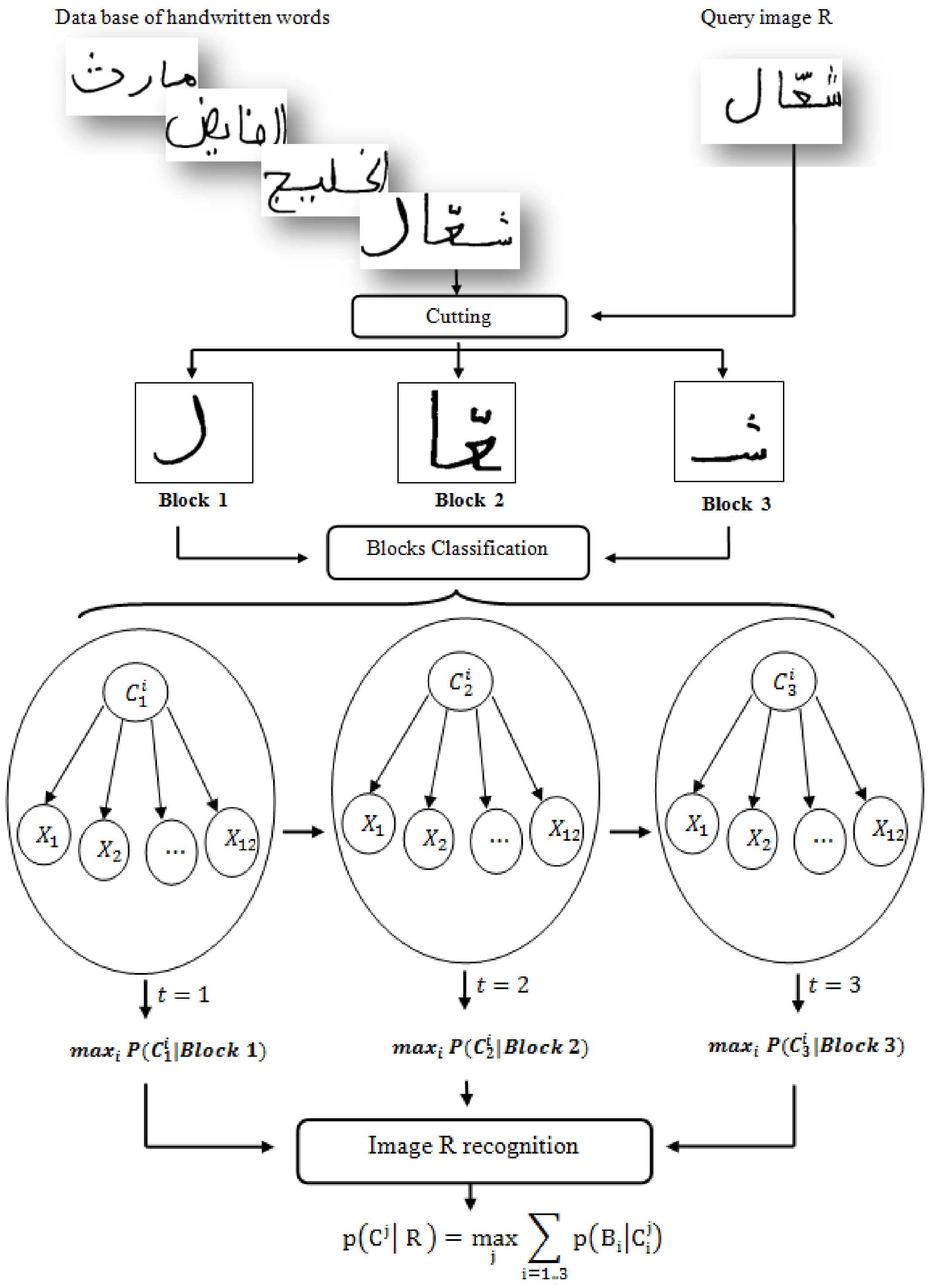

**Figure 3:** Proposed system structure

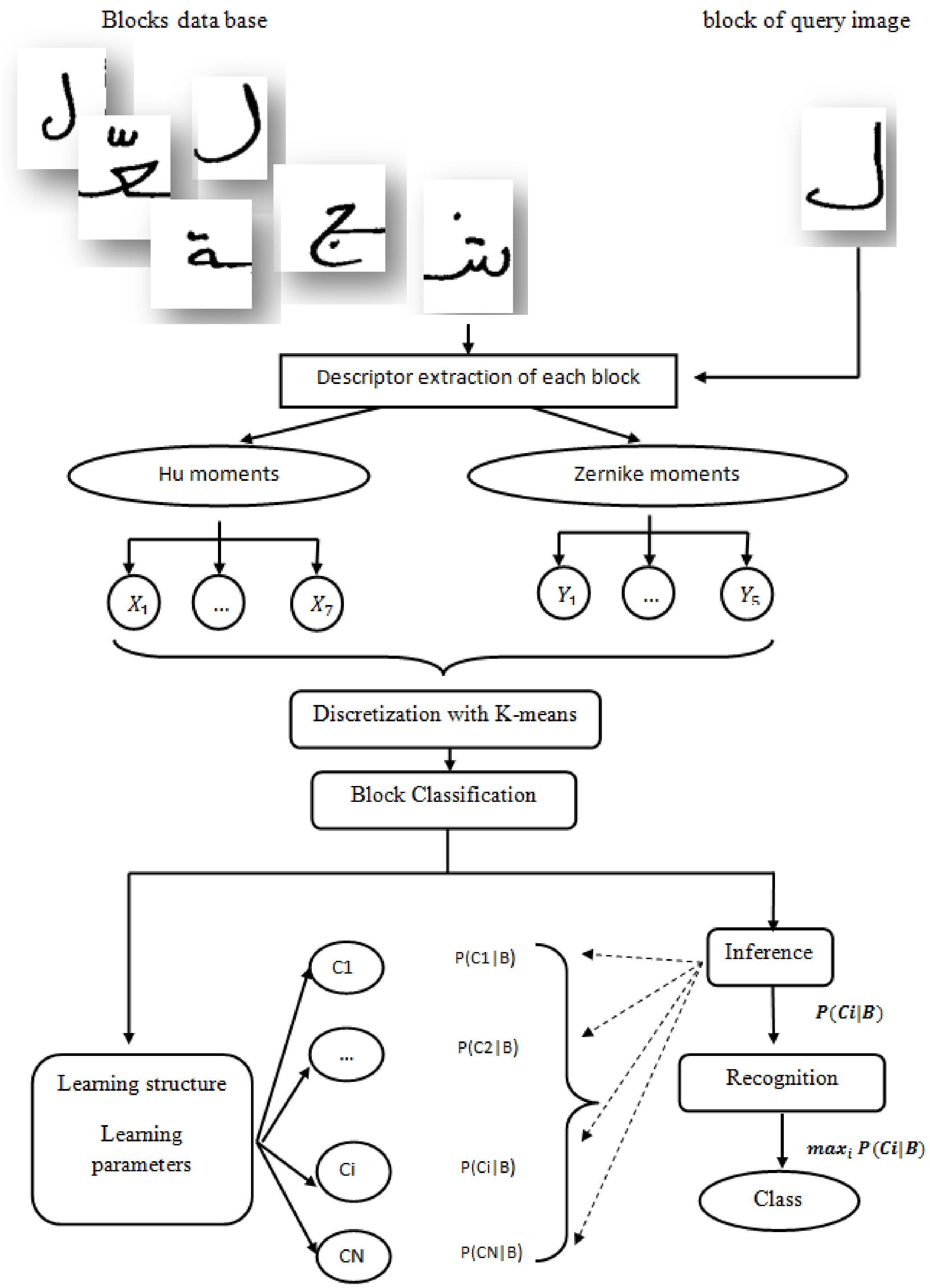

**Figure 4** : Proposed system structure

## 6. DYNAMIC BAYESIAN NETWORK

A dynamic Bayesian network (Suen et al., 1980) can be defined as a repetition of conventional networks in which we add a causal link (representing the time dependencies) from one time slice to another. The network in each time slice (figure 5) contains a number of random variables representing observations and hidden states of the process.

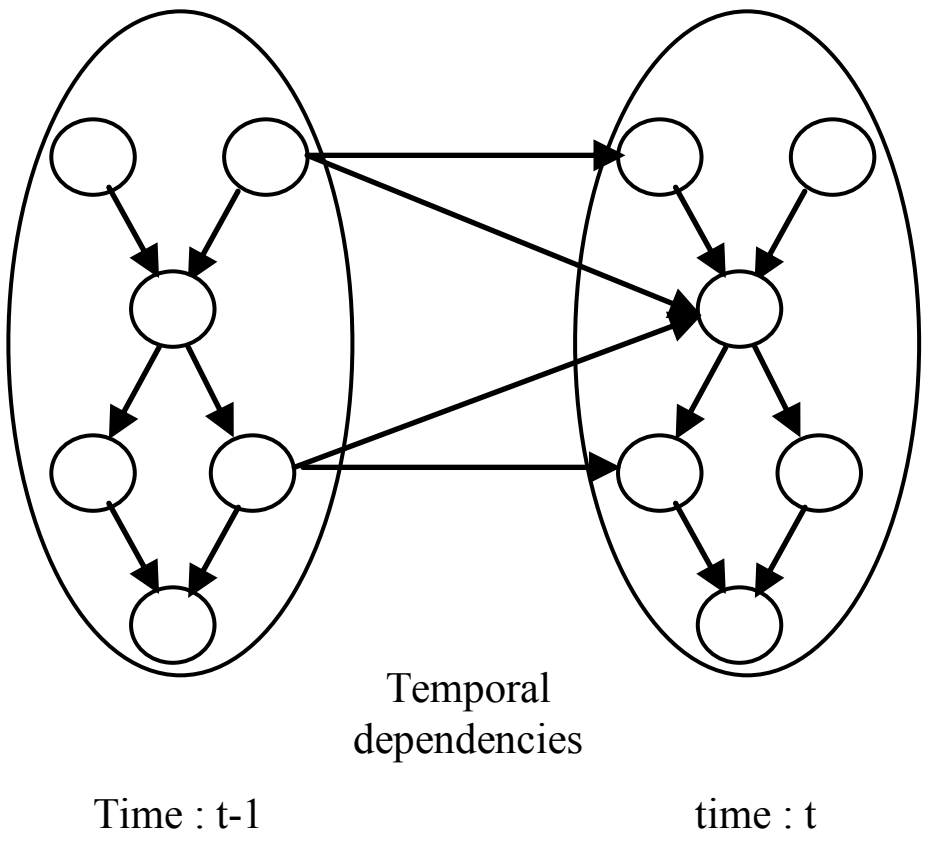

**Figure 5.** Representation of two slice times as DBN

Consider a dynamic Bayesian network composed of a sequence of T hidden state variables (a hidden state of DBN is represented by a set of hidden state variables) $X = \{x_1, \ldots, x_T\}$ and a sequence of T observable variables $Y = \{y_1, \ldots, y_T\}$ where T is the limit time of the studied process. In order to have a complete specification of this DBN, we need to define 1) transition probability between states $(x_t/x_{t-1})$ , The conditional probability of hidden states given an observation $P(y_t/x_t)$ and initial state probability $P(x_1)$

The first two parameters should be given at each time $t = 1, \ldots, T$. These parameters can be or not time invariant. A HMM can be viewed as specific case of the more general Dynamic Bayesian Network (Alper & Yarman-Vura, 1997) (figure 6):

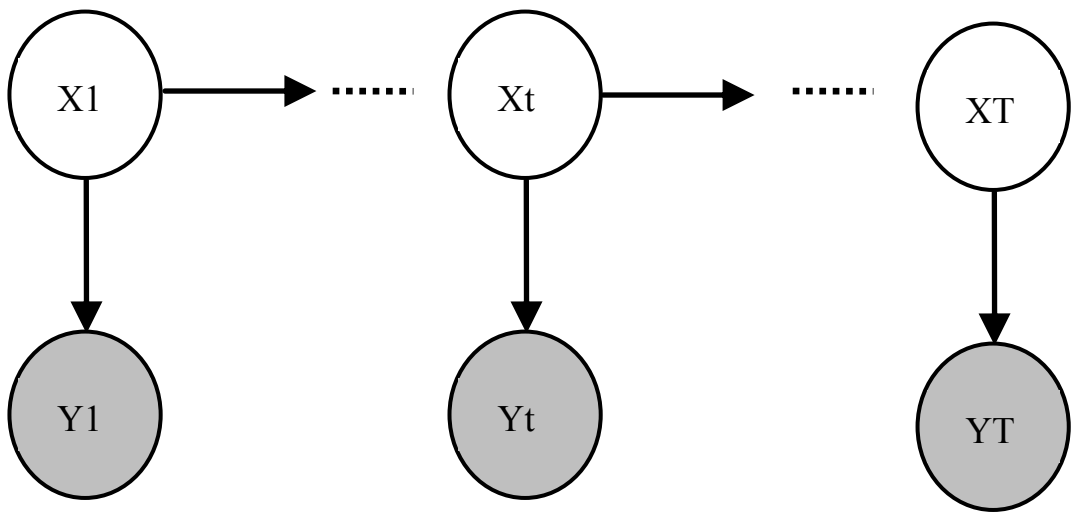

**Figure 6.** HMM represented as a DBN unrolled for 3 time slice

### 6.1 Modeling

Instead of considering a single (vertical or horizontal) HMM to model a given image of any handwritten word, we construct a more complex DBN by coupling the two HMMs. This coupling is performed by adding directed links between nodes in the graph to represent dependencies between state variables. But, the problem is how those links will be added.

#### 6.1.1 Graphical structure of the DBN model

The best way to determine the graphical structure of the model is to learn the structure from data. This technique, called structural learning (Miled, 1998), is not the goal of our approach. As part of our work, we will fix a graphical structure of DBN for all words images (figure 7). The choice of structure should be based on the following criteria:

- The chosen model structure must have a reasonable number of parameters in order to have an affordable computational complexity.
- No continuous variable should have discrete child in order to apply the exact inference algorithm (junction tree).
- Some arcs between the hidden variables of both HMMs must exist to model dependencies.

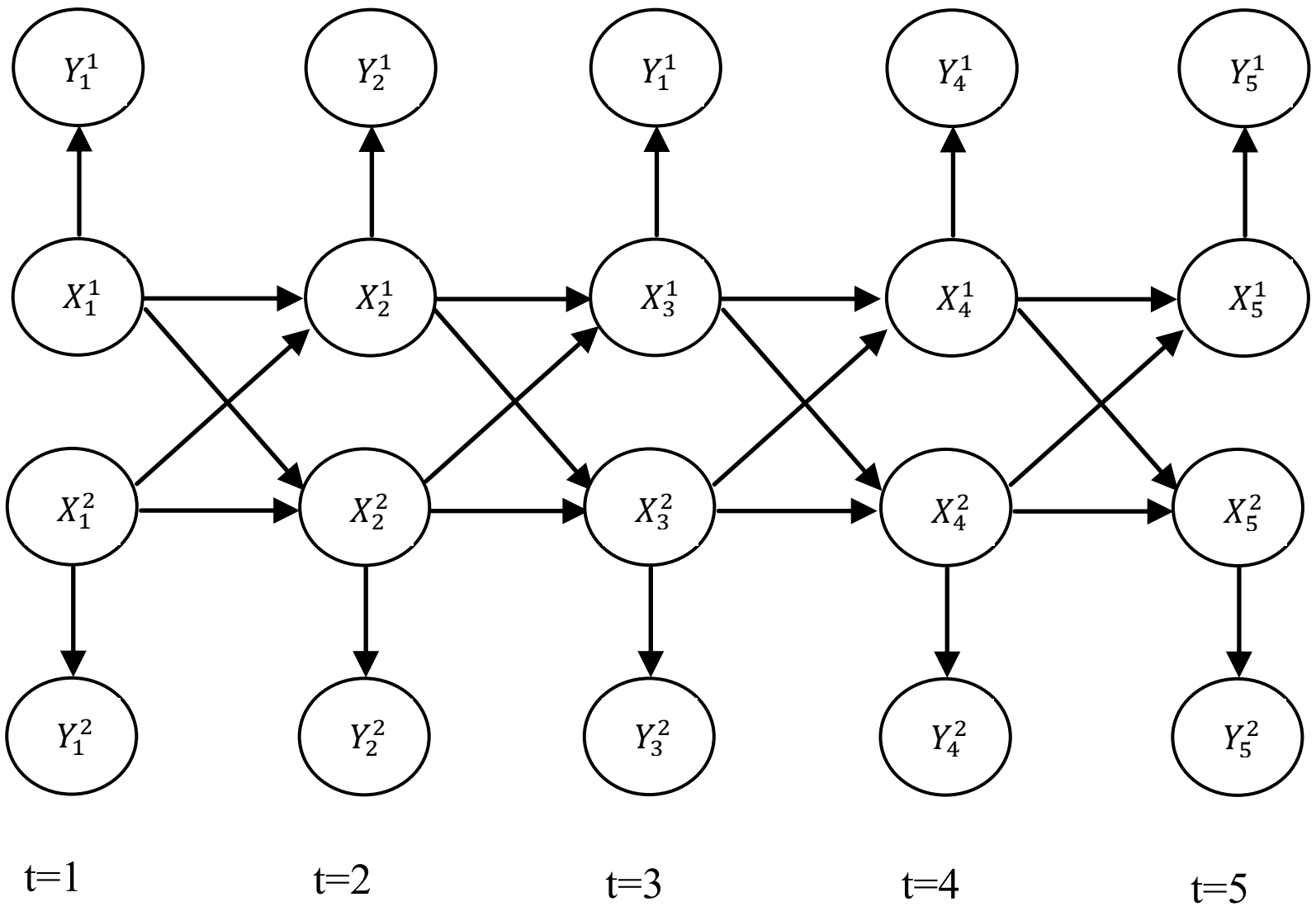

**Figure 7.** DBN model based on the coupling of two hidden Markov chains

In the above model, the dependencies between two HMMs modeling both horizontally and vertically data flow are performed by the relations between states, leading to efficient models in terms of complexity of the model (Pechwitz & Margner, 2003): a state of a HMM is connected to the adjacent state in the next time slice of the other HMM. We assume that the process modeled by the DBN is a first order Markov stationary process. In practice, this means that parents of any variable $X_t^i$ or $Y_t^i$ belong only to the time slices t and t-1 and the model parameters are independent of t. Therefore, the DBN model can then be represented by the first two time slots. It is fully described by giving the following parameters:

$$\begin{cases} \pi = \{\pi_i^{(l)}\} = P(X_1^l = i), \qquad l = 1,2 \\ A\,\{a_{i,j,k}^{(l)}\} = P(X_t^l = k \,/\, X_{t-1}^1 = i, \qquad X_{t-1}^2 = j), \qquad l = 1,2\ et\ t \geq 2 \\ B = \{b_{j,k}^{(l)}\} = P(Y_t^l = k / X_t^l = j), \ l = 1,2\ et\ t \geq 2 \end{cases}$$

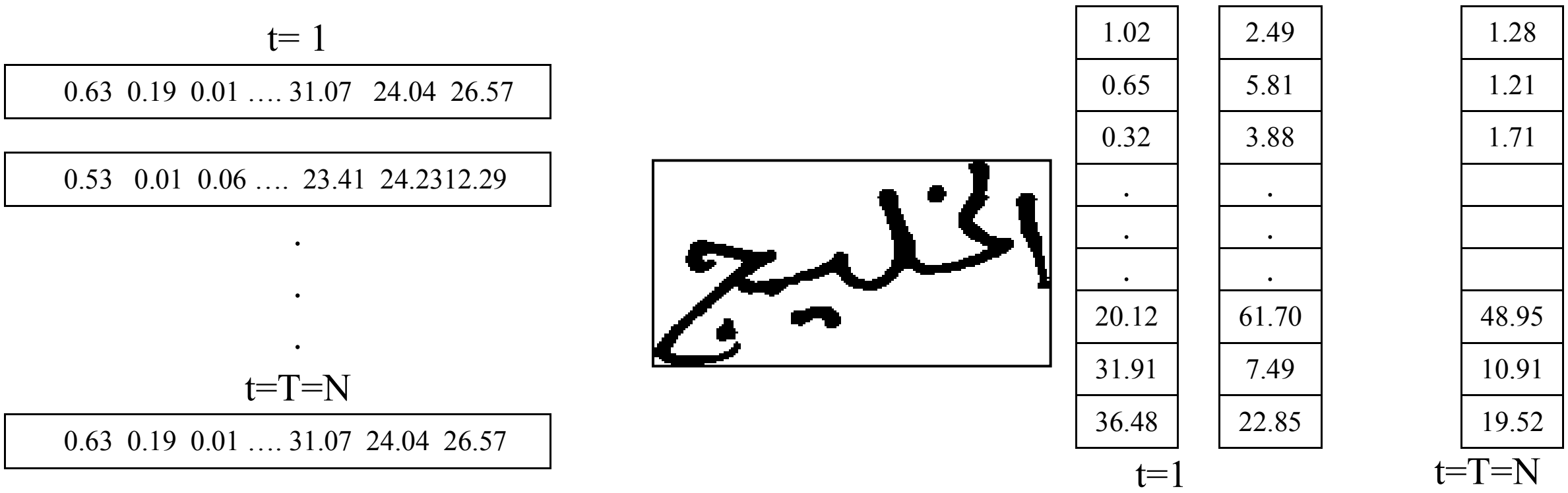

Figure 8. Example of features (7 Hu moments and 6 Zernike moments) calculated on an image extracted from the IFN / ENIT base.

### 6.1.2 Preprocessing and feature extraction of a handwritten word image

Each image of a handwritten word is transformed into two sequences of features vectors that will be the observations to be given to the DBN model: the first vectors features sequence models the flow of observations on the columns and is calculated using a sliding window moving horizontally from right to left. The second vectors feature sequence models the flow of observations on the rows and is calculated from a sliding window moving vertically from top to down (Figure 8).In our case, the analysis window size is uniform. No character segmentation procedure is made in a priori. Each word is converted into two matrices whose number of lines corresponds to the number of windows and the number of columns corresponds to the number of features. The calculated characteristics on each analysis window are the same as section 5.2 concerning naïve bayes (Hu and Zernike moments). In our approach, we hypothesized that the features extracted by scanning (horizontally or vertically) image of a word can be modeled by a discrete HMM. For this we use the k-means algorithm to discretize the obtained observations.

### 6.2 EM Algorithm for parameters learning

In the present work, we developed a model for each class. Models of all classes share a single DBN structure, but their parameters change from one class to another. Learning the model parameters is performed independently model by model, using the EM algorithm (Expectation Maximisation) (Suen, 1973). The EM algorithm is an iterative approach of maximum likelihood estimators. Each iteration of the EM algorithm is composed of two main steps: a step of expectation (E) and a step of maximizing (M). This algorithm aims to maximize the log likelihood function $l(\theta, \mathrm{Y}) = \log(L(\theta, \mathrm{Y}))$ where θ represents the set of model parameters and Y is the set of observations data relating to the problem. To apply the EM algorithm for learning our DBN model, we should note that this is a case of incomplete data. In this case, since the set of variables X is hidden, $l(\theta, \mathrm{X}, \mathrm{Y})$ becomes a random variable and can't be maximized directly. So the EM algorithm is based on maximizing the expectation of the distribution of X, the objective function, given the observations Y and the current parameters θ, using the auxiliary function : $Q(\theta, \hat{\theta}) = E_X[\log(L(\theta, X, Y))/\, Y, \hat{\theta}]$

1) In Expectation step (E), we start from the set of parameters θ and we calculate the auxiliary function : $Q(\theta, \hat{\theta}) = E_X[\log(L(\theta, X, Y))/\, Y, \hat{\theta}]$. If the hidden variables are continuous, then $Q(\theta, \hat{\theta}) = \int_{\mathrm{x} \in \mathrm{X}} \log(L(\theta, x, y)) \times L(x/y, \hat{\theta})\, dx$ If the hidden variables are discrete, then : $Q(\theta, \hat{\theta}) = \sum_{x \in X} \frac{L(x, y/\hat{\theta})}{L(y/\hat{\theta})} \times \log(L(x, y/\theta))$. The goal of expectation in a posteriori calculated by the previous function is to solve the problem of the presence of hidden variables variables X.

2) In maximisation step (M), we want to calcultae model parameters $\hat{\theta}_{new}$ : $\hat{\theta}_{\mathrm{new}} = \arg\max_{\theta} \mathrm{Q}(\theta; \hat{\theta})$ . This step allows maximizing the auxiliary function calculated in step (E) relative to the set of parameters.

### 6.3 general architecture of the proposed system

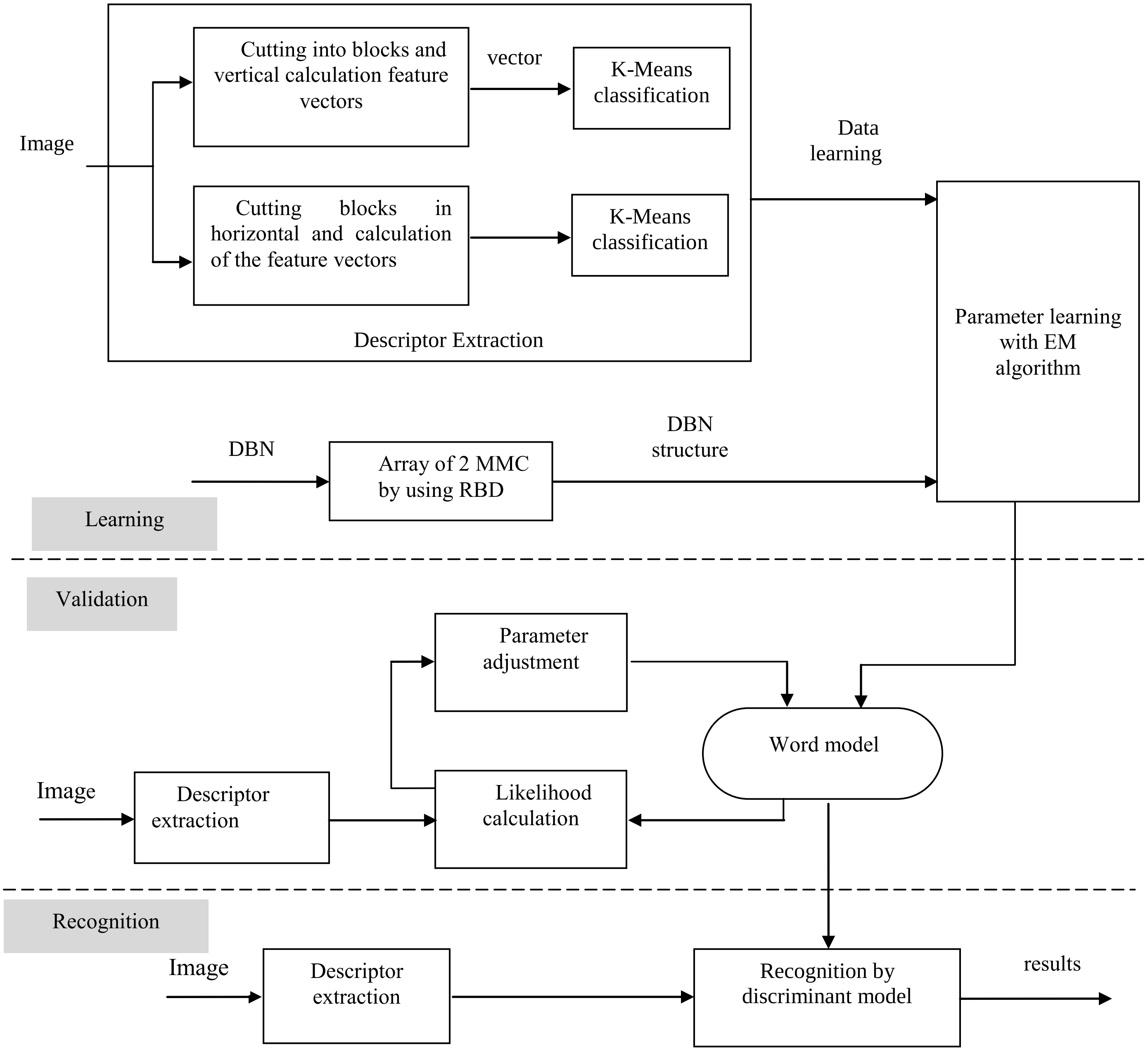

**Figure 9.** General architecture of the developped system

Overall, the proposed system is composed of three phases: a learning phase, a validation phase and a recognition phase. First, the image is divided into N vertical blocks and N horizontal blocks. A feature vector is calculated on each block. These vectors are discretized with the k-means method to obtain the data used to model the input of the DBN. The learning phase is used to generate the model of a given word from several sample images of the corresponding word by calculating the conditional probabilities associated with the used DBN. As for the validation phase, it adjusts the model parameters. During the recognition phase, the characteristics of the sample in question are extracted as in the learning phase (figure 9). Then, the likelihood of each model relative to the sample is calculated using an exact inference algorithm using junction tree algorithm and the word is assigned to the class that gives the maximum likelihood.

## 7 EXPERIMENTS AND RESULTS

Now, we present the results of the contribution of our approach to classify images of some examples of classes from the database used. The tests have been done on a corpus of Tunisian city names that is extracted from the IFN/ENIT database (Pechwitz et al., 2001). Our choice has been carried on city names of variable frequencies to show the capacity of the adopted approach to overcome the problem of lack of information. We have selected a subset of 18 models and generated a database of 3600 words, composed of 200 different images per model. Each word of city name corresponds to a model. On this database (table I) we have defined various training and test sets of different sizes

For each class, we established from the set of images in the base IFN / ENIT, three parts for learning, validation and test.

- Learning Base : 50% of the images of each class are used for learning the model parameters (conditional probability tables CPT)
- validation Database : 25% of the images of each class are used to adjust different parameters of the model
- Test Database : 25 % of the images of each class for testing the developed model.

We have used Matlab, and more exactly Bayes Net Toolbox of Murphy and Structure Learning Package to learn structure. Indeed, by applying architecture developed in section 5 and the algorithm of TAN and FAN detailed in our previous work (Jayech & Mahjoub, 2011), we obtained the figure 10.

| Class | City name | Images Example | Class | City name | Images Example |
|---|---|---|---|---|---|
| C1 | الرضاع | 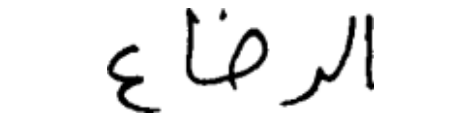 | C10 | المنزه 6 | 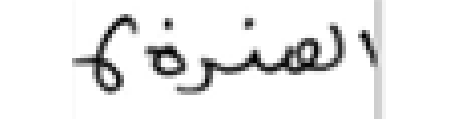 |
| C2 | شعال | 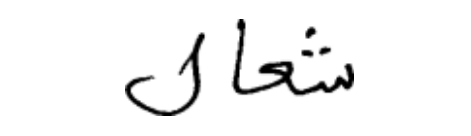 | C11 | النفيضة | 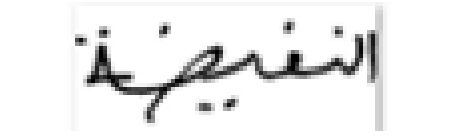 |
| C3 | نحال | 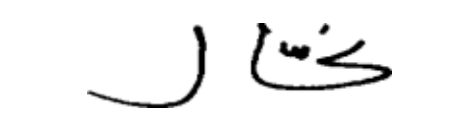 | C12 | نقة | 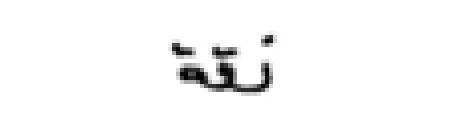 |
| C4 | مارث | 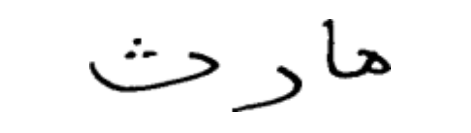 | C13 | الحامة | 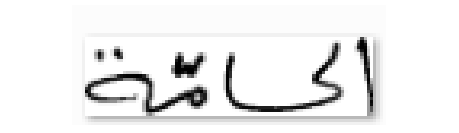 |
| C5 | شماخ | 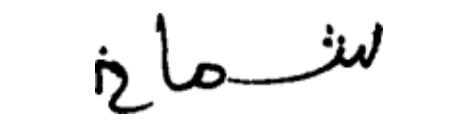 | C14 | عوام | 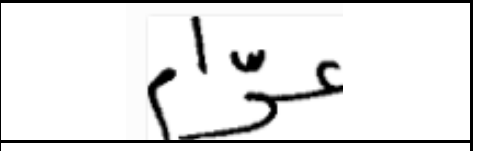 |
| C6 | الخليج | 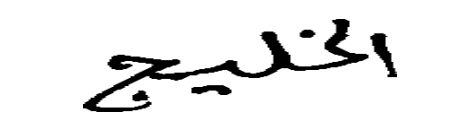 | C15 | رنوش | 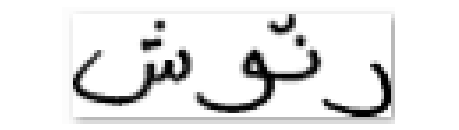 |
| C7 | الرقاب | 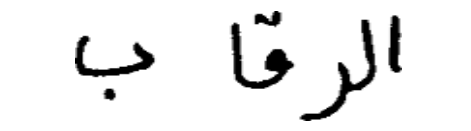 | C16 | بوزقام | 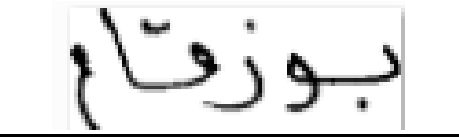 |
| C8 | الفايض | 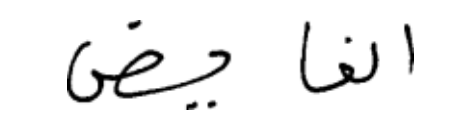 | C17 | خزندار | 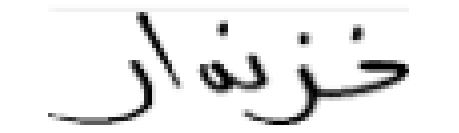 |
| C9 | سيدي إبراهيم الزهار | 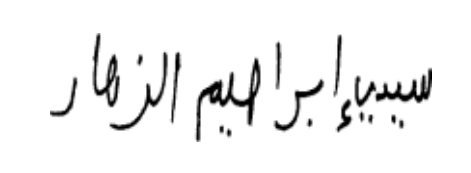 | C18 | المنصورة | 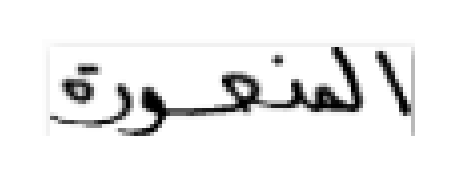 |

**Table 1.** Names of cities used for model experimentation

## 7.1 Naïve bayes

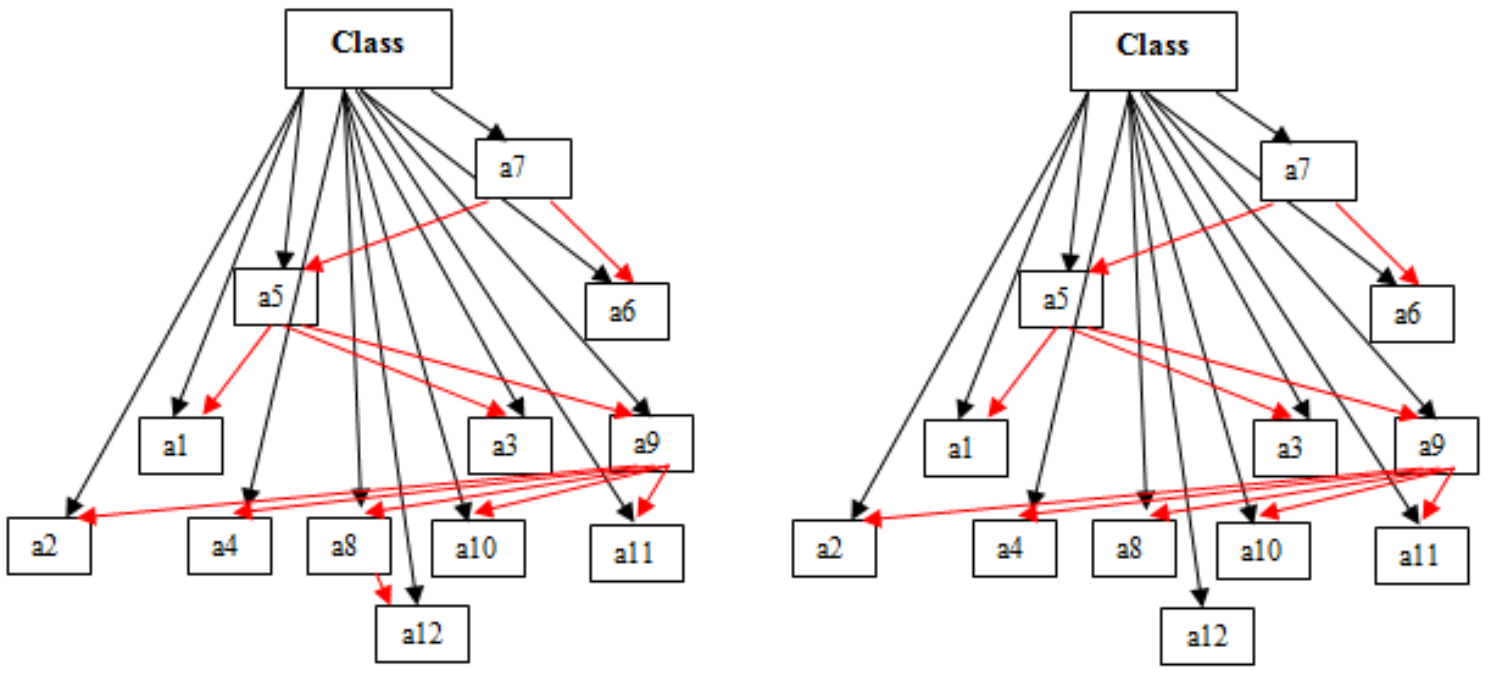

(a) (b)

**Figure 10.** Network structure (a) Structure of TAN ( b) Structure of FAN

We find that the rate of correct classification concerning the test data base for FAN (82.56%) which is better than obtained by TAN (80%) which is also better than obtained by NB (73%). According to our experiments, we observe that the naive Bayesian network gave a good result as TAN. Two factors may be the cause:

- The directions of links are crucial in a TAN. According to the TAN algorithm detailed in our article (Jayech K , Mahjoub M.A 2011) an attribute is randomly selected as the root of the tree and the directions of all links are made thereafter. We note that the selection of the root attribute actually determines the structure of the TAN result, since TAN is a directed graph. Thus the selection of the root attribute is important to build a TAN.
- Of unnecessary links can exist in a TAN (figure 10.a). According to the TAN algorithm, a spanning tree of maximum weight is constructed. Thus, the number of links is set to n-1. Sometimes it could be a possible bad fit of the data, since some links may be unnecessary to exist in the TAN.

It is observed that NB and Global Fan gave the same classification rate, since they have the same structure. We note also that the rate of correct classification given by FAN is very high that TAN. Several factors are involved; 1) According to the FAN algorithm illustrated in our article (Miled, 1998), the choice of the attribute A root is defined by the equation below, the maximum mutual root has the information with the class :$A_{root} = \mathrm{argmax}_{A_i} I_p(A_i; C) \quad \text{when } i = 1, \dots, n$.. It is obvious to use this strategy, ie the attribute that has the greatest influence on the classification should be the root of the tree 2) Filtering of links that have less than a conditional mutual information threshold. These links are at high risk for a possible bad fit of the data which could distort the calculation of conditional probabilities. Specifically, the use of a conditional average mutual information defined in the equation below as a threshold. All links that have conditional mutual information unless Iavg are removed.

$$Iavg = \frac{\sum_i \sum_{j, j \neq i} I_p(A_i; A_j | C)}{n(n-1)}$$ Where n is the number of attributes.

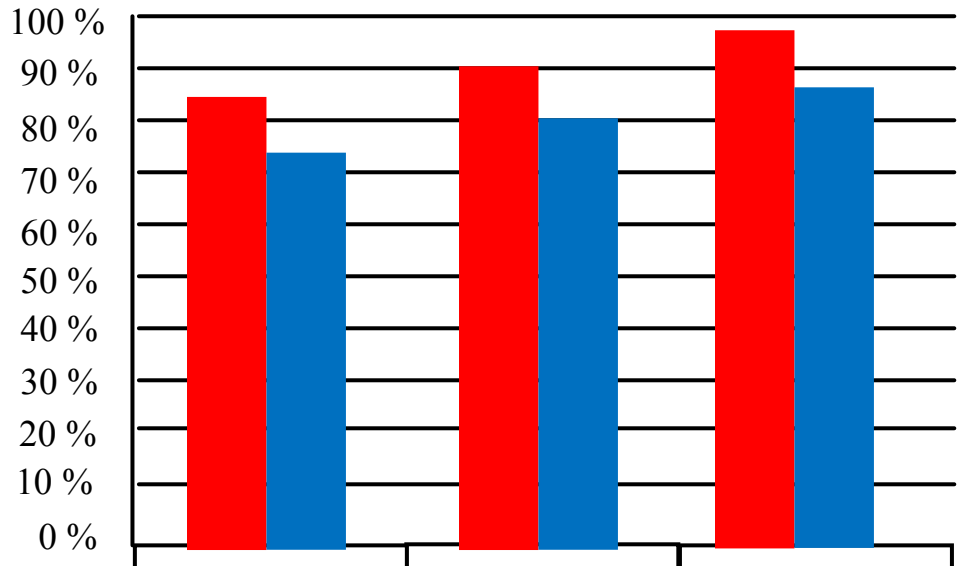

| | NB | TAN | FAN |
|---|---|---|---|
| Learning set | 85% | 90,66 | 93,50 |
| Test set | 73% | 80% | 82,56 |

**Figure 11.** Classification rates

### 7.2 Dynamic Bayesian netwok

The first experiments are conducted to find, using the validation dataset, the optimal number of states for the model for each class. The criteria used to determine the number of states is the recognition rate and the cost of inference in terms of time. To do this, we vary each time the number of states and determine the recognition rate and the cost of inference (figure 12).

All results concerning the optimal number of states for each model are illustrated in the following table:

| Class | C1 | C2 | C3 | C4 | C5 | C6 | C7 | C8 | C9 |
|---|---|---|---|---|---|---|---|---|---|
| Q | 15 | 19 | 14 | 15 | 14 | 11 | 15 | 17 | 14 |
| Class | C10 | C11 | C12 | C13 | C14 | C15 | C16 | C17 | C18 |
| Q | 15 | 15 | 15 | 16 | 19 | 12 | 15 | 16 | 116 |

**Table 2.** Optimum cardinality of the latent variable per class

After finding the optimal number of states for each class, we performed the recognition tests on samples of the test database. The following table shows the recognition rate generated by our system:

| Classe | C1 | C2 | C3 | C4 | C5 | C6 | C7 | C8 | C9 |
|---|---|---|---|---|---|---|---|---|---|
| Top1 | 85,8 | 82,2 | 81,9 | 80,4 | 80,9 | 86,2 | 85 | 84,7 | 87,4 |
| Classe | C10 | C11 | C12 | C13 | C14 | C15 | C16 | C17 | C18 |
| Top1 | 81.5 | 82,8 | 85,1 | 83,0 | 82,3 | 83.2 | 83 | 82,9 | 80,4 |

**Table 3.** Recognition rates per class

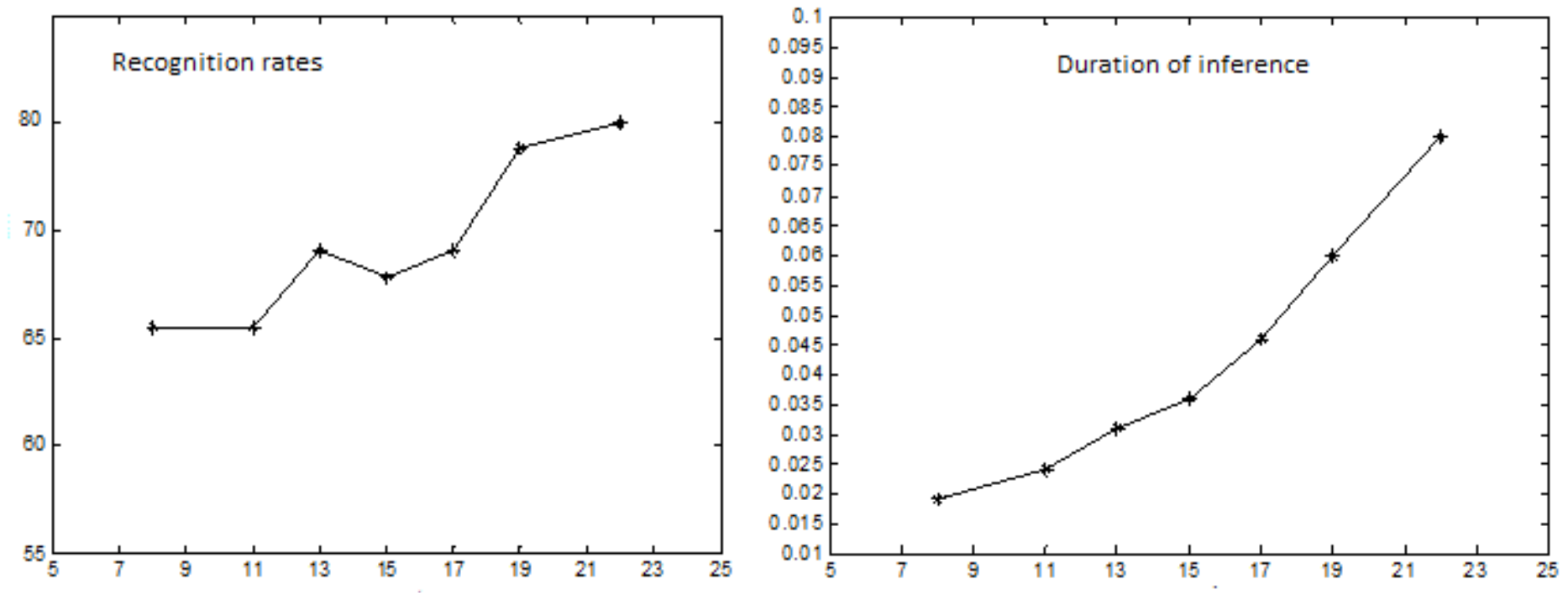

a) b)

**Figure 12. (a)** Curve illustrating the recognition rate depending on number of states for class C2, **(b)** Curve illustrating the duration of inference based on number of states for class C2

The overage rate recognition $T_m$ is defined as the average rate $T_i$ for each class $T_m = \frac{\sum_{i=1}^{n} T_i}{\mathrm{n}}$, where n is the number of class used for testing the model. The average rate recognition obtained using the test dataset is around 83.7% which is better than FAN. The following table shows the confusion rates on the test basis considering classes 1 to 9 :

| | **C1** | **C2** | **C3** | **C4** | **C5** | **C6** | **C7** | **C8** | **C9** |
|---|---|---|---|---|---|---|---|---|---|
| **C1** | 85,79 | | | | | 6,32 | 2,61 | 5,28 | |
| **C2** | | 82,22 | 4,42 | | 7,79 | | 5,57 | | |
| **C3** | | 6,89 | 77,49 | | 8,89 | | 6,73 | | |
| **C4** | | | | 80,45 | 6,81 | | 4,78 | 7,96 | |
| **C5** | | 3,95 | 8,85 | | 80,88 | | | 6,32 | |
| **C6** | 4,28 | | | | | 86,19 | | 9,53 | |
| **C7** | 1,25 | | | | | 5 | 85 | 8,75 | |
| **C8** | 2,23 | | | | | 5,64 | 8,43 | 84,7 | |
| **C9** | | 4,8 | 1,87 | | | 5,88 | | | 87,45 |

**Table 4.** Confusion rates on the test basis

Confusion rates are relatively high on this lexicon of 9 cities names. However, we note that the correct answer appears at most one of the first four system propositions.

## 8 CONCLUSION

In this paper, we presented a new approaches of the off-line Arabic handwriting recognition. We have used two type of Bayesian networks models as classifier to classify the whole image of handwritten Tunisian city names. We have developed a new approach for classifying images using the method of k-means to cluster the vectors descriptors of the images. We have implemented and compared four classifiers: Naïve Bayes (NB), TAN, FAN and DBN. The goal was to compare the results obtained by these structures and apply algorithms that can produce useful information from a high dimensional data. In particular, the aim is to improve Naïve Bayes by removing some of the unwarranted independence relations among features and hence we extend Naïve Bayes structure by implementing the Tree Augmented Naïve Bayes. Unfortunately, our experiments show that TAN performs even worse than Naïve Bayes in classification (figure 11). Responding to this problem, we have modified the traditional TAN learning algorithm by implementing a novel learning algorithm, called Forest Augmented Naïve Bayes. In addition, we note that the results recorded with DBN are better than those of FAN. On the other hand our best rates recorded (83.7 %) according to test base are compared to those found in the literature like illustrated in the following table:

| **Researches** | **Recognition rates** |
|---|---|
| S Touj 2007 (Touj et al., 2007) | 80.44% |
| R El hajji (El-Hajj et al., 2005) | 75-86% |
| F Menasri (2008) | 83-92% |
| M. Pechwitz (Pechwitz & Maergner, 2003) | 84% |
| J Al Khateeb (Al Khateeb, 2011) | 66% |

**Table 4:** Recognition rates of main researches of IFN data base

We believe that the results are promising and that many improvements can be made to the proposed model, especially regarding the descriptors calculated on blocks of the image, the distribution law of these descriptors and the initialization of learning algorithm.